\documentclass{article}
\usepackage{ijcai18}

\usepackage{times}
\usepackage{xcolor}
\usepackage{soul}
\usepackage[utf8]{inputenc}
\usepackage[small]{caption}

\usepackage{amssymb}
\usepackage{amsmath}
\usepackage{comment}
\usepackage{algorithm}
\usepackage{algorithmic}
\usepackage{amsmath}
\usepackage{moresize}
\usepackage{helvet}
\usepackage{courier}
\usepackage{graphicx}
\usepackage{wrapfig}
\usepackage{url}
\usepackage[font=small]{caption}
\usepackage{subcaption}

\usepackage{times}
\def\i#1{\hbox{\it #1\/}}
\def\beq{\begin{equation}}
\def\eeq#1{\label{#1}\end{equation}}
\def\ba{\begin{array}}
\def\ea{\end{array}}

\def\t{\textbf{t}}
\def\f{\textbf{f}}

\def\iif{\hbox{\bf if}}

\def\causes{\hbox{\bf causes}}
\def\inertial{\hbox{\bf inertial}}
\def\default{\hbox{\bf default}}

\def\nonex{\hbox{\bf nonexecutable}}

\usepackage[inline,ignoremode]{trackchanges}

\addeditor{lb}
\addeditor{yf}
\addeditor{dm}

\title{PEORL: Integrating Symbolic Planning and Hierarchical Reinforcement Learning for Robust Decision-Making}
\author{
Fangkai Yang$^1$, 
Daoming Lyu$^2$, 
Bo Liu$^2$,
Steven Gustafson$^1$
\\ 
$^1$ Maana Inc., Bellevue, WA, USA  \\
$^2$ Auburn University, Auburn, AL, USA\\
fyang@maana.io,
daoming.lyu@auburn.edu,
boliu@auburn.edu,
sgustafson@maana.io
}

\begin{document}

\maketitle

\begin{abstract}
Reinforcement learning and symbolic planning have both been used to build intelligent autonomous agents. Reinforcement learning relies on learning from interactions with real world, which often requires an unfeasibly large amount of experience. Symbolic planning relies on manually crafted symbolic knowledge, which may not be robust to domain uncertainties and changes. In this paper we present a unified framework {\em PEORL} that integrates symbolic planning with hierarchical reinforcement learning (HRL) to cope with decision-making in a dynamic environment with uncertainties.
 Symbolic plans are used to guide the agent's task execution and learning, and the learned experience is fed back to symbolic knowledge to improve planning. This method leads to rapid policy search and robust symbolic plans in complex domains. The framework is tested on benchmark domains of HRL. 
\end{abstract}
\section{Introduction}
Reinforcement learning (RL) \cite{sutton1998reinforcement} and symbolic planning \cite{cim08} have both been used to build autonomous agents that behave intelligently in the real world. An {\em RL agent} relies on interactions with the environment to achieve its optimal behavior, without the need of prior knowledge. Building upon the model of Markov Decision Process (MDP), the {\em policy}, i.e., a mapping from a state to an action, can be learned via a number of trial-and-error. 
With the recent development of deep learning, such methods can lead to a highly adaptive and robust agent \cite{mnih2015human}, but may often rely on an unfeasibly huge amount of experience. 
On the other hand, a large body of work on symbolic task planning of mobile robots exists~\cite{hanheide2015robot,chen2016planning,khandelwal2017bwibots}, where a {\em planning agent} carries prior knowledge of the dynamic system, represented in a formal, logic-based language such as PDDL \cite{mcdermott1998pddl} or an action language \cite{gel98} that relates to logic programming under answer set semantics (answer set programming) \cite{lif08}. The agent utilizes a symbolic planner, such as a PDDL planner \textsc{FastDownward}~\cite{helmert2006fast} or an answer set solver \textsc{Clingo}~\cite{gekasc12c} to generate a sequence of actions to achieve its goal.
The pre-defined, manually crafted symbolic representation may not completely capture all domain details, and domain uncertainties and execution failures are handled by execution monitoring and re-planning.
A planning agent does not require a large number of trial-and-error (which is quite expensive for real robots) to behave reasonably well, but it may not be robust enough to domain uncertainties, change and some reward structure that is only available through execution and learning. As planning and reinforcement learning are important and complementary aspects of intelligent behavior, combining the two paradigms to bring out the best of both worlds is quite attractive. In this paper, we focus on developing such a framework where an agent
(i) utilizes a symbolic representation to generate plans that guide reinforcement learning
, and 
(ii) leverages learned experience to enrich symbolic knowledge and improve planning.

Topics (i) and (ii) above have been studied separately by researchers from different communities. (i) has been studied in a broader sense by the RL community based on the notion of hierarchical reinforcement learning (\textit{HRL}) \cite{Barto:2003:RAH:608557.608576}. The HRL framework has a two-level structure: the lower level is called the (primitive) action level, with the primitive actions as defined in the MDP setting, and the higher level is called the task level, or termed as the {\em option} level~\cite{hrl:option:sutton1999}. An option is a temporal abstraction of actions specified by a policy and a termination condition, 
and guides learning such that at any state, only the primitive actions specified by the option are considered.
Symbolic plans play similar roles \cite{Ryan02usingabstract,leonetti2016synthesis}, but these work does not embrace the hierarchical aspect of options: symbolic actions are 1-1 corresponding to primitive actions in the underlying MDP. Nor do they dynamically discover new plans and options to improve learning. For (ii), basic learning models such as relational decision trees \cite{COIN:COIN447} or weighted exponential average \cite{kha14} were used to improve planning through execution experiences, but they are not as expressive or general as a general RL framework.  

Aiming at building an agent that can unify planning and RL for robust decision-making, this paper advances both lines of research by integrating symbolic planning using action language ${\cal BC}$ \cite{lee13} with hierarchical R-learning \cite{rlearning:schwartz,sm:mlj96} through a {\em Planning--Execution--Observation--Reinforcement-Learning} (PEORL) framework.
R-learning is an important family of reinforcement learning paradigm that characterizes finite horizon average reward, and is shown to be particularly suitable for planning and scheduling tasks.
In PEORL, we use ${\cal BC}$ to represent commonsense knowledge of actions and constraint answer set solver {\sc Clingcon} \cite{Banbara2017ClingconTN} to generate a symbolic plan, given an initial state and a goal. The symbolic plan is then mapped to a deterministic sequence of stochastic options to guide RL. R-learning iterates on two values: the average-adjusted reward $R$ and the cumulative average reward, or termed as the gain reward $\rho$. 
While $R$-values indicate the learned policy, $\rho$-values can be effectively used by {\sc Clingcon} to generate an improved symbolic plan with better quality, in terms of the cumulative gain reward. The improved plan is mapped to new options, which further guide R-learning to continue, until no better symbolic plan can be found. 
The framework is empirically evaluated on two benchmarks: Taxi domain \cite{Barto:2003:RAH:608557.608576} and Gridworld \cite{leonetti2016synthesis}. In our experiments, when the algorithm terminates, the optimal symbolic plan is returned.

The contribution of this paper is summarized as follows:
\begin{itemize}
\item To advance learning capability of agents, to the best of our knowledge, this is the first work using symbolic planning for option discovery in HRL. The PEORL agent outperforms RL agent and HRL agent by returning policies of significantly larger cumulative reward.
\item To advance planning capability of agents, to the best of our knowledge, this is the first work where symbolic planning leverages R-learning to improve its robustness. The PEORL agent outperforms planning agent by discovering a new state that leads to extra reward and reducing the number of execution failure.
\end{itemize}
The paper is organized as follows. After a brief review of action language ${\cal BC}$ and HRL in Section~\ref{sec:prelim}, we present the framework formulation in Section~\ref{sec:formulation} and the main algorithm in Section~\ref{sec:learning}. Experimental evaluation results are shown in Section~\ref{sec:exp}. 
Related work is discussed in Section~\ref{sec:related}.

\section{Preliminaries}\label{sec:prelim}

\textbf{Action Language and Symbolic Planning.} An {\em action description} $D$ in the language ${\cal BC}$ \cite{lee13} includes two kinds of symbols, {\em fluent constants} that represent the properties of the world, denoted as $\sigma_F(D)$, and {\em action constants}, denoted as $\sigma_A(D)$. A fluent atom is an expression of the form $f=v$, where $f$ is a fluent constant and $v$ is an element of its domain. For boolean domain, denote $f=\t$ as $f$ and~$f=\f$ as $\sim\!\!\! f$. An action description is a finite set of {\em causal laws} that describe how fluent atoms are related with each other in a single time step, or how their values are changed from one step to another, possibly by executing actions. For instance,
$
(A~\iif~A_1,\ldots,A_m)
$
is a {\em static law} that states at a time step, if $A_1,\ldots, A_m$ holds then $A$ is true. Another static law
$
(\default~f=v)
$
states that by default, the value of $f$ equals~$v$ at any time step. 
$
(a~\causes~A_0~\iif~A_1,\ldots, A_m)
$
is a {\em dynamic law}, stating that at any time step, if $A_1,\ldots, A_m$ holds, by executing action $a$,~$A_0$ holds in the next step. 
$
(\nonex~a~\iif~A_1,\ldots,A_m)
$
states that at any step, if $A_1,\ldots, A_m$ holds, action $a$ is not executable. Finally, the dynamic law
$
(\inertial~f)
$
states that by default, the value of fluent $f$ does not change from one step to another, formalizing the {\em commonsense law of inertia} that addresses the frame problem.

An action description captures a dynamic transition system. A {\em state} $s$ is a complete set of fluent atoms, and a transition is a tuple $\langle s_1,a,s_2 \rangle$ where $s_1, s_2$ are states and~$a$ is a (possibly empty) set of actions. The semantics of $D$ is defined by a translation into a set of answer set programs~$\i{PN}_l(D)$, for an integer~$l\ge 0$ stating the maximal steps of transition. It is shown that all answer sets of $\i{PN}_0(D)$ correspond to all states in the transition system, and all answer sets of $\i{PN}_l(D)$ correspond to all transition paths $\Pi$ of length $l$, of the form $\langle s_1,a_1,\ldots, a_{l-1}, s_l \rangle$ (or equivalently, $\Pi=\bigcup^{l-1}_1\langle s_i, a_i, s_{i+1}\rangle$) \cite[Theorems 1, 2]{lee13}. Let $I$ and $G$ be states. The triple $(I,G,D)$ is called a planning problem. $(I,G,D)$ has a plan of length $l-1$ iff there exists a transition path of length $l$ such that $I=s_1$ and $G=s_l$. Throughout the paper, we use $\Pi$ to denote both the plan and the transition path by following the plan.

Due to the semantic definition above, automated planning with an action description in ${\cal BC}$ can be achieved by an answer set solver, and the output answer sets encode the transition paths that solve the planning problem. 

\noindent\textbf{R-learning for Average Reward.} A Markov Decision Process (MDP)   is defined as a tuple $({\mathcal{S},\mathcal{A},P_{ss'}^{a},r,\gamma})$, where $\mathcal{S}$ and $\mathcal{A}$ are the sets of symbols denoting states and actions, the transition kernel $P_{ss'}^{a}$ specifies the probability of transition from state $s\in\mathcal{S}$ to state $s'\in\mathcal{S}$ by taking action $a\in\mathcal{A}$, $r(s,a):\mathcal{S}\times\mathcal{A}\mapsto\mathbb{R}$ is a reward function bounded by $r_{\max}$, and $0\leq\gamma<1$ is a discount factor. A solution to an MDP is a policy $\pi:\mathcal{S}\mapsto \mathcal{A}$ that maps a state to an action. RL concerns on learning a near-optimal policy  by executing actions and observing the state transitions and rewards, and it can be applied even when the underlying MDP is not explicitly given, a.k.a, model-free policy learning.

To evaluate a policy $\pi$, there are two types of performance measures: the expected discounted sum of reward for infinite horizon problems 
and the expected un-discounted sum of reward for finite horizon problems. In this paper we adopt the latter metric defined as $J^\pi_{\rm avg}(s) = \mathbb{E}[\sum\limits_{t = 0}^T {{r_t}}|s_0=s ]$. We define the \textit{gain reward} ${\rho ^\pi }(s)$ reaped by policy $\pi$ from $s$ as
$$
{\small
{\rho ^\pi }(s) = \mathop {\lim }\limits_{T \to \infty } \frac{{J^\pi_{{\rm{avg}}}(s)}}{T} = \mathop {\lim }\limits_{T \to \infty } \frac{1}{T}\mathbb{E}[\sum\limits_{t = 0}^T {{r_t}} ]
} .
$$ 
R-learning \cite{rlearning:schwartz,sm:mlj96} 
 is a model-free value iteration algorithm that can be used to find the optimal policy for average reward  criteria. At the $t$-th iteration $(s_t,a_t,r_t, s_{t+1})$, update:
$$
\ba{rl}
 {R_{t + 1}}({s_t},{a_t})\!\! & \xleftarrow{\alpha_t} {r_t} - {\rho _t}(s_t)  
 + \mathop {\max }\limits_a {R_t}({s_{t + 1}},a)\\
 \rho_{t+1}(s_t)\!\!& \xleftarrow{\beta_t} r_t + \mathop {\max }\limits_a {R_t}({s_{t + 1}},a) - \mathop {\max }\limits_a {R_t}({s_t},a),
\ea
$$
where $\alpha_t, \beta_t$ are the learning rates, and $a_{t+1} \xleftarrow{\alpha} b$ denotes the update law as ${a_{t + 1}} = (1-\alpha){a_{t}} + \alpha b$.


\noindent\textbf{Hierarchical Reinforcement Learning.} Compared with regular RL, hierarchical reinforcement learning (HRL) \cite{Barto:2003:RAH:608557.608576} specifies on real-time-efficient decision-making problems over a series of tasks.
An MDP can be considered as a flat decision-making system where the decision is made at each time step. On the contrary, humans make decisions by incorporating temporal abstractions.
An option is temporally extended course of action consisting of three components: a policy $\pi: {\mathcal{S}} \times {\mathcal{ A }} \mapsto {{ [0, 1]}}$, a termination condition~$\beta: {\mathcal{S}} \mapsto {{ [0, 1]}}$, and an initiation set ${\mathcal{I}} \subseteq {\mathcal{S}}$. An option~$(I,\pi ,\beta)$ is available in state $s_t$ iff ${s_t} \in I$. After the option is taken, a course of actions is selected according to~$\pi$ until the option is terminated stochastically according to the termination condition $\beta$.
With the introduction of options, the decision-making has a hierarchical structure with two levels, where the upper level is the option level (also termed as task level) and the lower level is the (primitive) action level. Markovian property exists among different options at the option level. 


\section{PEORL Framework}\label{sec:formulation}
In this section we formally define PEORL framework. A PEORL theory is a tuple~$(I, G, D,\mathcal{S}, \mathcal{A}, r, \gamma, \mathbb{F}_A)$. It contains the elements from a symbolic planning problem, an MDP and how they are linked with each other:
\begin{itemize}
\item $I, G, D$ form a symbolic planning problem, where $I$ is the initial state, $G$ is a {\em PEORL goal} that consists of a goal state condition and a linear constraint, and $D$ is a {\em PEORL action description} in the language of ${\cal BC}$. 
\item $\mathcal{S}, \mathcal{A},r,\gamma$ form part of an MDP. $\mathcal{A}$ is a set of action symbols in MDP space. We use small letters with tilde, such as $\tilde{a}$ to denote its element, and assume $|\sigma_A(D)|\le|\mathcal{A}|$.~$\mathcal{S}$ is a set of state symbols in MDP space. It contains {\em simple state} symbols of form $s$ which are 1-1 correspondent to (symbolic) states of~$T(D)$. Due to such correspondence, we also use a state of $T(D)$, i.e., a set of fluent atoms in~$\sigma_F(D)$ to denote a simple state symbol in $\mathcal{S}$. Furthermore, $\mathcal{S}$ also contains the {\em MDP state symbols}, denoting a state obtained by applying MDP action $\tilde{a}$. $r$ is a reward function such that $r(s,a):\mathcal{S}\times\mathcal{A}\mapsto \mathbb{R}$. $0\le \gamma\le 1$ is a discount factor.
\item A {\em symbolic transition--option} mapping $\mathbb{F}_A$ that translates a symbolic transition path $\Pi\subseteq T(D)$ into a set of options.
\end{itemize}
Some components are further explained as follows.
\subsection{Symbolic Planning Problem}
A {\em PEORL action description} $D$ is written in ${\cal BC}$ and contains a specific set of causal laws formulating {\em plan quality} accrued from executing a course of actions:
\begin{itemize}
\item For any state of $T(D)$ that contains atoms $\{A_1,\ldots, A_n\}$, $D$ contains static laws of the form 
\beq
 s~\iif~A_1,\ldots, A_n,~\hbox{for~simple~state}~s\in\mathcal{S}.
\eeq{srule}
\item Introduce new fluent symbols of the form $\rho(s,a)$ to denote the gain reward at state $s$ following action $a$. $D$ contains a static law stating by default, the gain reward is a sufficiently large number, denoted as~$\i{INF}$, to promote exploration when necessary:
$$
\!\!\!\default~\rho(s,a)=\i{INF},~\hbox{for simple state}~s\in\mathcal{S},a\in\sigma_A(D).
$$
\item Use fluent symbol $\i{quality}$ to denote the cumulative gain reward reward of a plan, termed as {\em plan quality}. $D$ contains dynamic laws of the form
\beq
\!a~\causes~\i{quality}=C+Z~\iif~s,\rho(s,a)=Z,\i{quality}=C.\!
\eeq{cost}  
\item $D$ contains a (possibly empty) set $P$ of facts of the form $\rho(s,a)=z$. 
\end{itemize}

A {\em PEORL initial state} contains a state $I$. In particular, the initial plan quality is 0. A {\em PEORL goal} $G=(A,L)$ where $A$ is a goal state, and $L$ is a linear constraint of the form
$
(\i{quality}\ge n)
$
where $n$ is an integer. The negation of $L$ is defined in the usual way.

The triple $(I, (A,L), D)$ forms a symbolic planning problem with linear constraints: the plan is encoded by a transition path of $T(D)$ that starts from state $I$ and ends in state $A$ with $L$ satisfied. A plan $\Pi$ of $(I, G, D)$ is {\em optimal} iff $\sum_{\langle s,a,t\rangle\in\Pi} r(s,a)$ is maximal among all plans. However, note that reward function $r$ is not a part of the planning problem because we assume that reward is part of the specific domain details not captured as prior knowledge. We later use PEORL learning algorithm to interact with the environment and generate optimal plan when the algorithm terminates.

Solving the planning problem follows the method of translating $I$, $G$ and $D$ into the input language of \textsc{Clingo} \cite{kha14}. However, in this paper, we use a slightly different but equivalent translation into the input language of \textsc{Clingcon} to handle linear constraints more efficiently\footnote{To use \textsc{Clingcon}, (\ref{cost}) is translated to
$$
\ba{l}
{\tt
\&sum\{quality(k-1);Z\}=quality(k):-}\\
~~~~~~~~~~~~~~~~~{\tt a(k-1),s(k-1),R(s,a,Z).}
\ea
$$
where {\tt k} stands for time step.
}.

\subsection{From Symbolic Transitions to Options}
\begin{figure*}[!btp]
\begin{subfigure}{.5\textwidth}
\centering
    \includegraphics[width=0.9\linewidth]{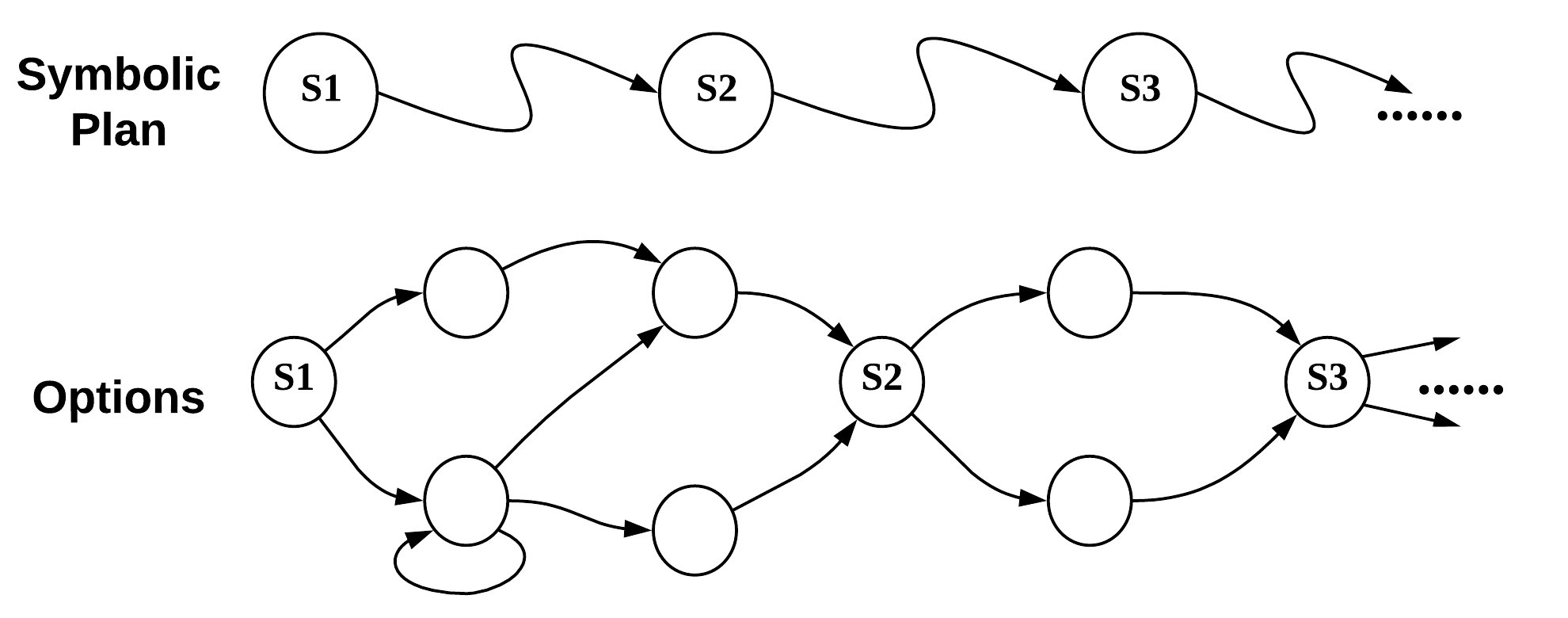}
  \subcaption{The mapping from a symbolic transition path to options}
  \label{fig:option}
 \end{subfigure}
 \begin{subfigure}{.5\textwidth}
  \centering
    \includegraphics[width=1.\linewidth]{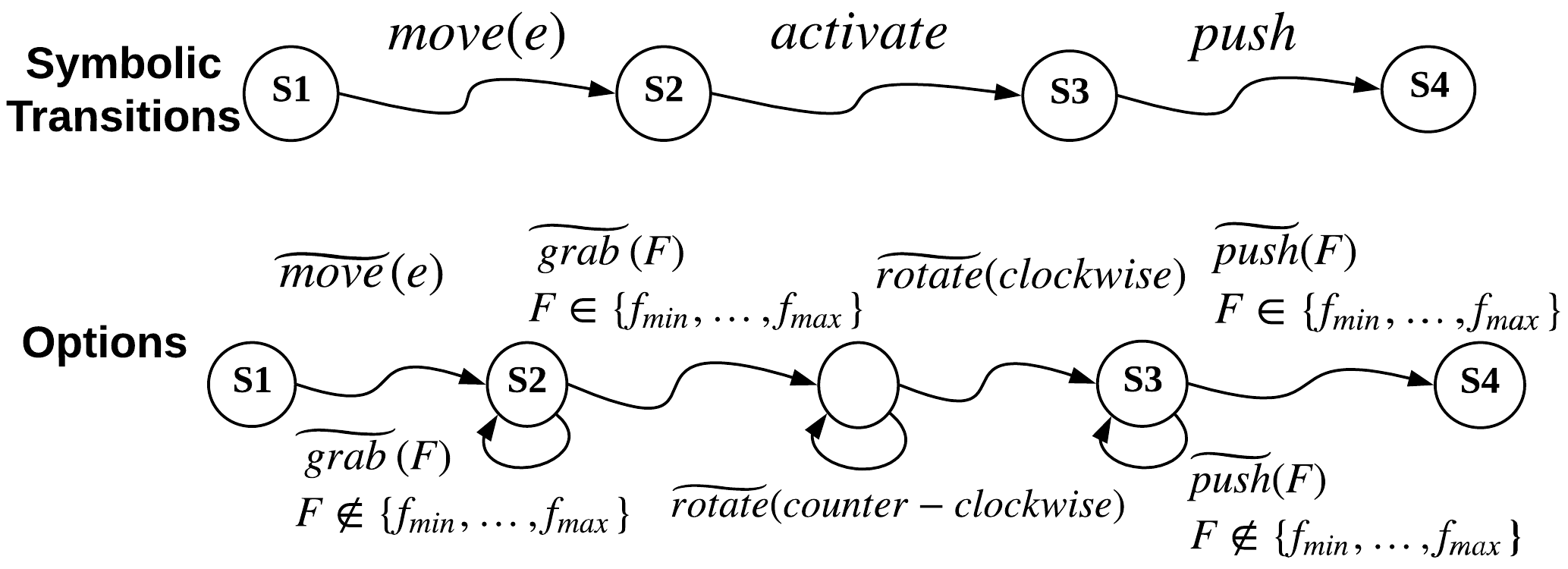}
  \subcaption{The option mapping for transitions $t_1,t_2,t_3$}
  \label{fig:gridworldmapping}
\end{subfigure}
\caption{Mappings from symbolic transitions to options}
\end{figure*}
We assume that in the transition system $T(D)$, for each transition $\langle s,a,t\rangle\in T(D)$, $a$ contains exactly one action symbol, i.e, concurrent execution of actions is not allowed. $\mathbb{F}^A$ maps a symbolic transition $\langle s,a,t\rangle$ to an option in the sense of \cite{Barto:2003:RAH:608557.608576}. $\mathbb{F}^A(\langle s,t,a\rangle) = (\pi,\beta,\mathcal{I})$ where $\pi:\mathcal{S}\times\mathcal{A}\mapsto [0,1]$, $\beta:\mathcal{S}\mapsto [0,1]$, and $\mathcal{I}\subseteq \mathcal{S}$. In particular, we enforce that option $\mathbb{F}^A(\langle s,t,a\rangle)$ is available for transition $\langle s,a,t\rangle$ iff
$
s=\mathcal{I}~~\hbox{and}~~\beta(t)=1.
$ 
This condition guarantees that the right option is chosen to realize the symbolic transition $\langle s,a,t\rangle$ at its starting state, and terminates when it fulfills the symbolic transition. 

We further build one more deterministic layer by mapping a transition path defined by a symbolic plan to a set of options.
For a transition path $\Pi=\langle s_1,a_1,\ldots, a_{l-1}, s_l\rangle$,
$${\small\mathbb{F}_A(\Pi)=\bigcup_{\langle s_{i-1},a_{i-1},s_i\rangle\in\Pi}\mathbb{F}^A(\langle s_{i-1},a_{i-1},s_i\rangle).}$$
It is easy to see that the execution of a symbolic plan is deterministically realized by executing their corresponding options sequentially. Such hierarchical mapping is illustrated in Figure~\ref{fig:option}. 

\subsection{Example: Grid World}\label{sec:gridworld}
We use the Grid World adapted from \cite{leonetti2016synthesis} as an example. In a $20\times 20$ grid, there is an agent that needs to navigate to (9,10), which can only be entered through (9,9). At (9,9) there is a door that the agent needs to activate first, and then push to enter. The action description consists of causal laws formulating effects of $\i{move}(E)$ where $E\in\{e,s,w,n\}$, $\i{push}$ and $\i{activate}$, for instance,
$$
\ba{lr}
\i{move}(e)~\causes~\i{pos}(X,Y+1)~\iif~pos(X,Y)\\
\nonex~\i{move}(e)~\iif~\i{pos}(X,20)\\
\nonex~\i{move}(e)~\iif~\i{pos}(9,9),\sim\!\!\i{dooropen}\\
\i{activate}~\causes~\i{dooractive}~\iif~pos(9,9),\sim\!\!\i{dooractive}\\
\i{push}~\causes~\i{dooropen}~\iif~pos(9,10),\i{dooractive}.
\ea
$$
Declare the following fluents are inertial:
$$
\inertial~\i{pos}~~~~\inertial~\i{dooropen}~~~~\inertial~\i{dooractive}.
$$
The following causal laws are instantiation of (\ref{srule}) and (\ref{cost}). They formulate the effects on plan quality by executing $\i{move}$ for a particular state, and similar causal laws can be defined for {\em activate} and {\em push}:
$$
\ba{l}
s(X,Y)~\iif~\i{pos}(X,Y),\sim\!\!\i{dooractive},\sim\!\!\i{dooropen}\\
\i{move}(E)~\causes~\i{quality}=C+Z~\iif~\\
~~~~~~~~~~~~s(X,Y),\rho(s(X,Y),\i{move}(E))=Z,\i{quality}=C.
\ea
$$
Assuming initially the agent is located at $(9,8)$ with door closed and inactive, the action description $D$, initial state $I=\{\i{pos}(9,8),\sim\!\!\i{dooractive},\sim\!\!\i{dooropen}\}$ and goal state $G=\{\i{pos}(9,10),\i{dooractive},\i{dooropen}\}$ are translated into the input language of \textsc{Clingcon} and a plan is
\beq
\!\!\!\!\!\ba{ll}
t_1: &\!\!\!\langle \{\i{pos}(9,8),\sim\!\!\i{dooractive},\sim\!\!\i{dooropen}\}, \i{move}(e),\\

&~~~~~\{\i{pos}(9,9),\sim\!\!\i{dooractive},\sim\!\!\i{dooropen}\}\rangle\\
t_2: &\!\!\!\langle \{\i{pos}(9,9),\sim\!\!\i{dooractive}, \sim\!\!\i{dooropen}\}, \i{activate},\\

&~~~~~\{\i{pos}(9,9),\i{dooractive},\sim\!\!\i{dooropen}\}\rangle\\
t_3: &\!\!\!\langle \{\i{pos}(9,9),\i{dooractive}, \sim\!\!\i{dooropen}\}, \i{push},\\

&~~~~~\{\i{pos}(9,9),\i{dooractive},\i{dooropen}\}\rangle\\
t_4: &\!\!\!\langle \{\i{pos}(9,9),\i{dooractive},\i{dooropen}\}, \i{move}(e),\\
&~~~~~\{\i{pos}(9,10),\i{dooractive},\i{dooropen}\}\rangle.
\ea
\eeq{path}
Now we map symbolic transitions~$t_1,t_2,t_3$ to options. As options talk about the realization of symbolic actions in terms of MDP actions, we assume that each symbolic action $\i{move}(E)$ for a direction $E$ is executed in the same way in MDP, denoted as $\widetilde{move}(E)$. Symbolic action $\i{push}$ can be executed in a variety of ways: the agent needs to use proper force to push the door such that the door can be opened without any damage. Therefore, $\i{push}$ is executed in finite number of options, denoted as $\widetilde{\i{push}}(F)$ where $F\in \{f_{min},\ldots,f_{max}\}$. Executing symbolic action {\em activate} as an option involves two steps: first, the agent needs to grab the doorknob using proper force, denoted by $\widetilde{\i{grab}}(F)$, where $F\in \{f_{min},\ldots,f_{max}\}$. Second, after the door knob is successfully grabbed, it can be turned either clockwise or counter-clockwise, and turning it clockwise can activate the door. This action is denoted as $\widetilde{\i{rotate}}(E)$ for $E\in\{\i{closewise},\i{counter-clockwise}\}$. The mapping from~$t_1,t_2,t_3$ to options is demonstrated as Figure~\ref{fig:gridworldmapping}.

\section{PEORL Learning}\label{sec:learning}
Given any transition $\langle s_{i-1},a_{i-1},s_{i}\rangle$ in a plan $\Pi$, hierarchical R-learning involves the updates of $R$ and $\rho$ in two steps. Since every symbolic transition is 1-1 correspondence to its option $\mathbb{F}^A(\langle s_{i-1},a_{i-1},s_i\rangle)$, we also use $a_{i-1}$ to denote the option. Before an option terminates, execute actions following the option, and for any transition $\langle x,\tilde{a},y\rangle$ where $\tilde{a}\in\mathcal{A}$, update
\beq
\ba{l}
R_{t+1}(x,\tilde{a})\xleftarrow{\alpha} r(x,\tilde{a})-\rho_t^{\tilde{a}}(x) + \max_{\tilde{a}} R_t(y,\tilde{a})\\
\rho^{\tilde{a}}_{t+1}(x)\xleftarrow{\beta}r(x,\tilde{a})+\max_{\tilde{a}} R_t(y,\tilde{a}) - \max_{\tilde{a}} R_t(x,\tilde{a}).
\ea
\eeq{riter0}
When option terminates, update
\beq
\!\!\!\ba{l}
R_{t+1}(s_{i-1},a_{i-1})\xleftarrow{\alpha} r(s_{i-1},a_{i-1}) - \rho_t^{a_{i-1}}(s_{i-1})\\
~~~~~~~~+\max_{a} R(s_i,a)\\
\rho_{t+1}^{a_{i-1}}(s_{i-1})\xleftarrow{\beta}r(s_{i-1},a_{i-1})\\
~~~~~~~~+\max_{a} R_t(s_i,a) - \max_{a} R_t(s_{i-1},a) ,
\ea
\eeq{riter}
where $\alpha$ and $\beta$ are learning rates for $R$ and $\rho$, $r$ denotes the cumulative reward accrued by executing option mapped from symbolic action $a_{i-1}$.
Given a plan $\Pi$, the quality of $\Pi$ is defined by summing up all gain rewards for the transitions in $\Pi$:
\beq
\i{quality}_t(\Pi) = \sum_{\langle s_{i-1},a_{i-1},s_i\rangle\in\Pi} \rho^{a_{i-1}}_t(s_{i-1}) .
\eeq{quality}
\begin{algorithm}
{\small
  \caption{PEORL Learning Loop}
  \label{algexec}
  \begin{algorithmic}[1]
    \REQUIRE $(I,G,D,\mathbb{F}_A)$ where $G=(A,\emptyset)$, and an exploration probability $\epsilon$
    \STATE $P_0\Leftarrow \emptyset$, $\Pi\Leftarrow \emptyset$
    \WHILE{True}
      \STATE $\Pi_o\Leftarrow \Pi$
      \STATE take $\epsilon$ probability to solve planning problem and obtain a plan $\Pi\Leftarrow\textsc{Clingcon}.\i{solve}(I,G,D\cup P_t)$
      \IF {$\Pi=\emptyset$}
          \RETURN $\Pi_o$
      \ENDIF
      \FOR {$\langle s_{i-1}, a_{i-1}, s_i\rangle\in\Pi$}
      \STATE use option $\mathbb{F}^A(\langle s_{i-1}, a_{i-1}, s_i\rangle)$ to update $R$ and $\rho$ by (\ref{riter0}) until the option terminates
      \STATE update $R$ and $\rho$ using (\ref{riter}).
      \ENDFOR
      \STATE calculate quality of $\Pi$ by (\ref{quality}).
      \STATE update planning goal $G\Leftarrow (A, \i{quality}> \i{quality}_t(\Pi))$.
      \STATE update facts $P_t\Leftarrow \{\rho(s_{i-1},a_{i-1})=z:\langle s_{i-1},a_{i-1},s_i\rangle\in\Pi, \rho_t^{a_{i-1}}(s_{i-1})=z\}$
    \ENDWHILE
  \end{algorithmic}}
\end{algorithm}
Given a PEORL theory $(I,G,D,\mathbb{F}_A)$, its learning algorithm is shown in Algorithm~\ref{algexec}. 

While previous results show that R-learning converges, most properties of option-based hierarchical R-learning remain unknown and therefore it remains an open question that option-based hierarchical R-learning converges to optimal over a finite number of options. For this reason, we leave the theoretical study of Algorithm~\ref{algexec} for our future work, but will empirically show its effectiveness on two benchmark domains in next section.

\begin{figure*}[htb!]
  \begin{subfigure}{.14\textwidth}
  \centering
  \includegraphics[width=.90\linewidth]{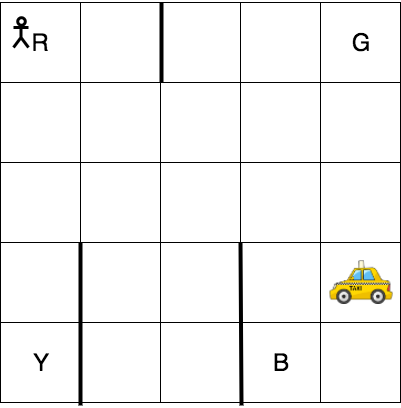}
  \caption{Taxi domain}
  \label{fig:taxi-exp}
\end{subfigure} 
  \begin{subfigure}{.14\textwidth}
  \centering
  \includegraphics[width=.90\linewidth]{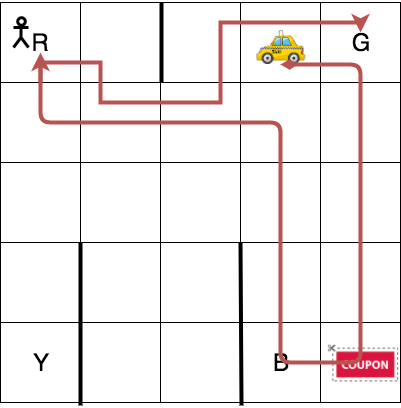}
  \caption{{\footnotesize A solution}}
  \label{fig:taxi-route}
\end{subfigure}
\begin{subfigure}{.36\textwidth}
  \centering
  \includegraphics[width=1\linewidth]{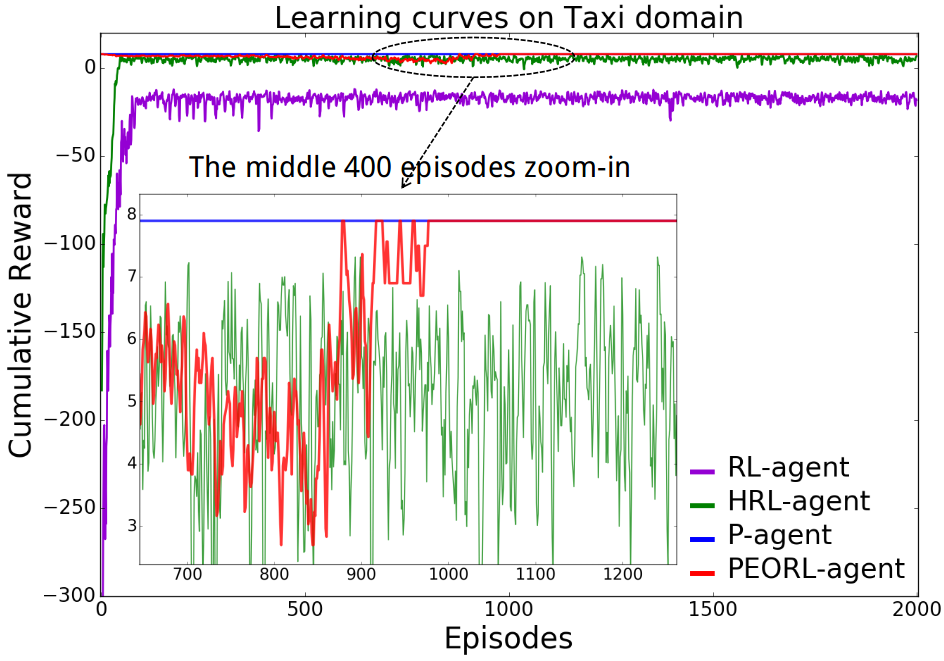}
  \caption{Learning curves on Taxi domain}
  \label{fig:taxi-reward}
\end{subfigure}
\begin{subfigure}{.36\textwidth}
  \centering
  \includegraphics[width=1\linewidth]{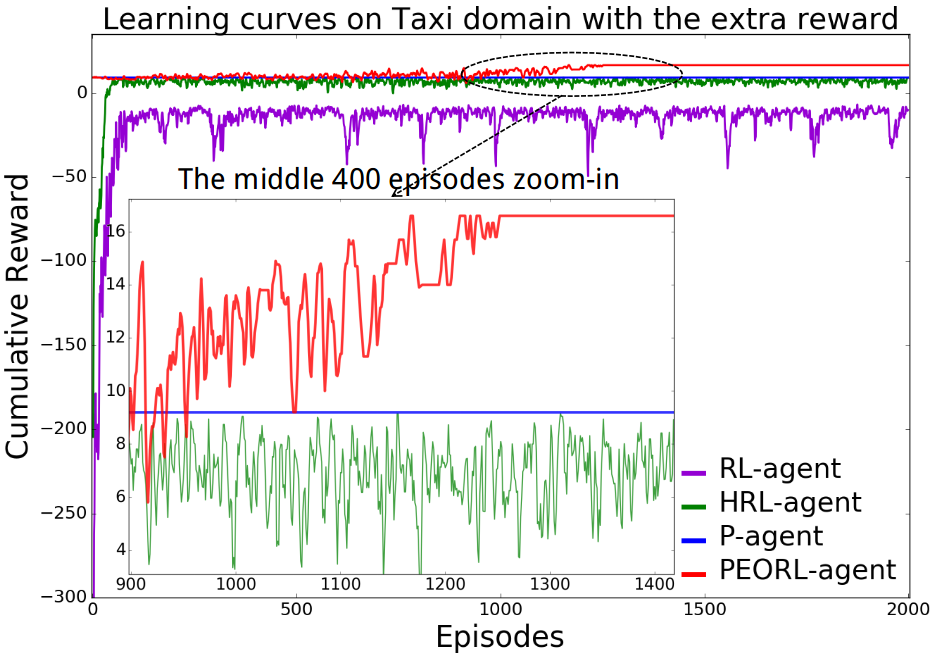}
  \caption{Learning curves with the extra reward}
  \label{fig:taxi-extra-reward}
\end{subfigure}
\caption{Taxi domain}
\end{figure*}

\begin{figure*}[htb!]
\begin{subfigure}{.14\textwidth}
  \centering
  \includegraphics[width=.90\linewidth]{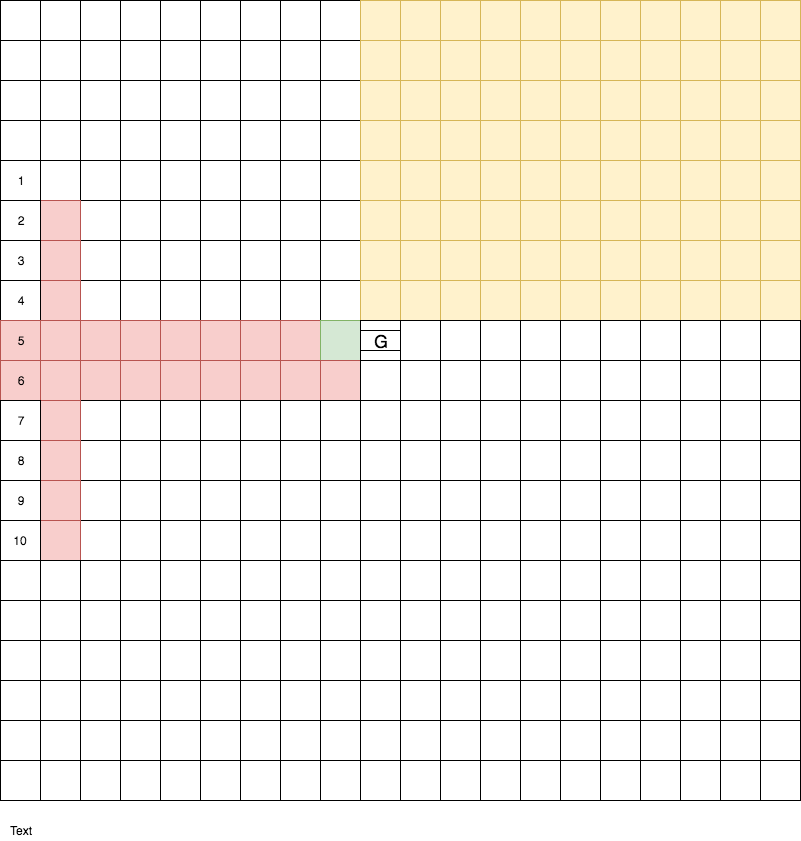}
  \caption{Grid World}
  \label{fig:gridworld-exp}
\end{subfigure}
\begin{subfigure}{.14\textwidth}
  \centering
  \includegraphics[width=.90\linewidth]{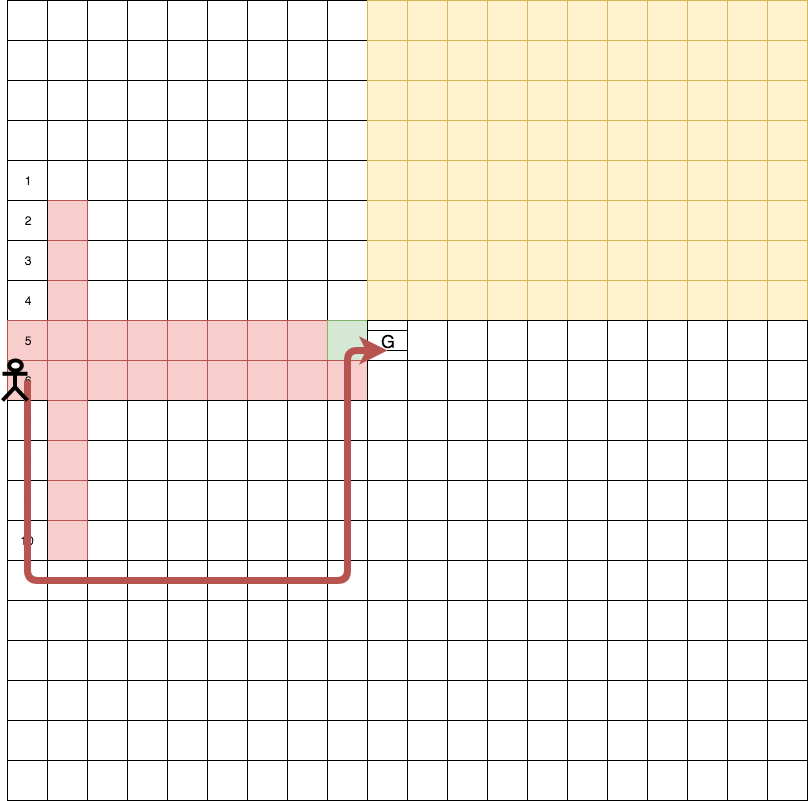}
  \caption{{\footnotesize A solution}}
  \label{fig:gridworld-exp-route}
\end{subfigure}
\begin{subfigure}{.36\textwidth}
  \centering
  \includegraphics[width=1\linewidth]{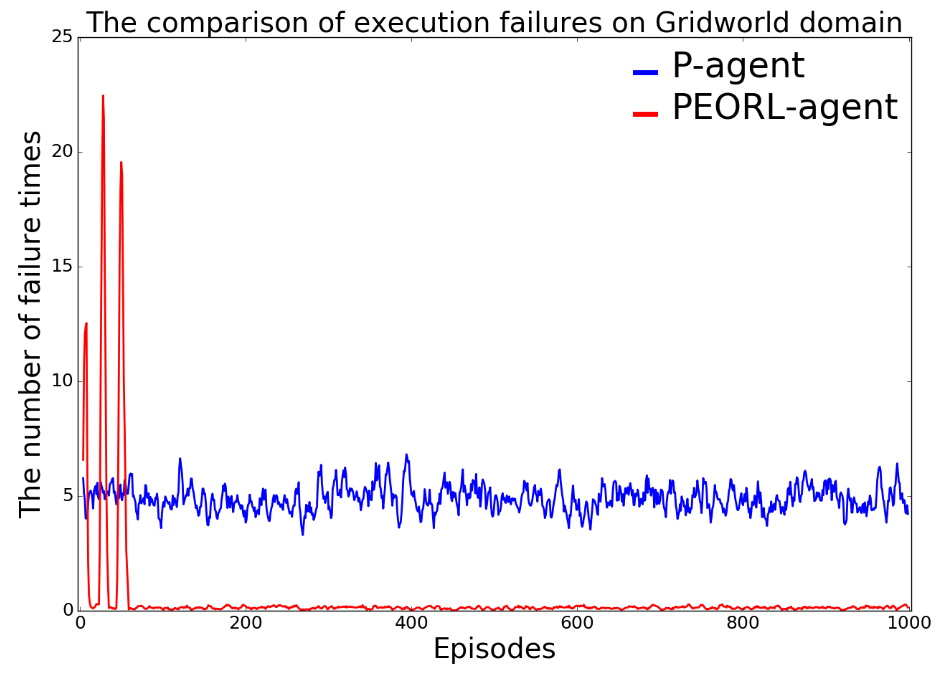}
  \caption{{\small Execution failure of PEORL and Planning}}
  \label{fig:gridworld-failures}
\end{subfigure}
\begin{subfigure}{.36\textwidth}
  \centering
  \includegraphics[width=1\linewidth]{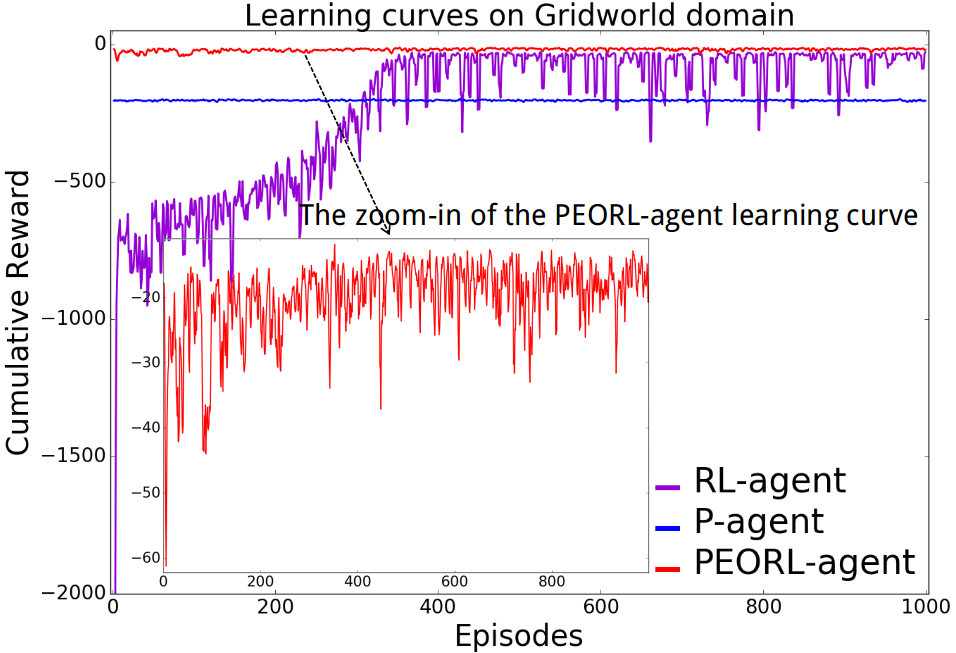}
  \caption{Learning curves on Gridworld domain}
  \label{fig:gridworld-reward}
\end{subfigure}
\caption{Grid World}
\end{figure*}

\section{Experiment}\label{sec:exp}

\subsection{Taxi Domain}

We first use Taxi domain \cite{Barto:2003:RAH:608557.608576} which is a benchmark domain for studying HRL. Scenario 1 is based on Taxi-v1 in OpenAI Gym (\url{https://gym.openai.com/envs/Taxi-v1/}). 
A Taxi starts at any location in a 5$\times 5$ grid map (Figure~\ref{fig:gridworld-exp}), navigates to a passenger, picks up the passenger, navigates to the destination and drops off the passenger, with randomly chosen locations for passenger and destination from marked grids. 
Every movement has a reward -1. Successful drop-off receives reward 20. Improper pick-up or drop-off receive reward -10. All actions are deterministic and always successful. In our experiment, we randomly set $10$ initial configurations and compare the cumulative rewards received by a standard Q-learning RL-agent, a HRL-agent based on hierarchical Q-learning using the manually crafted options specified in \cite{Barto:2003:RAH:608557.608576}, a standard planning agent (P-agent) using \textsc{Clingo} to generate plans and execute, and a PEORL-agent. For all learning rates $\alpha$ is annealed from 1 to 0.01, and for PEORL agent, $\beta=0.5$. The result (Figure~\ref{fig:taxi-reward}) shows the cumulative reward of PEORL agent significantly surpasses the RL-agent and is also superior to HRL-agent. Guided by its symbolic plan, PEORL agent has a clear motivation to achieve its goal. For this reason, it never commits actions that violate its commonsense knowledge, such as an improper pick-up or drop-off, or run into the walls. For this reason, the penalty of -10 never occurs to PEORL agent, so the variance of the cumulative reward is a lot smaller than RL-agent and HRL-agent. PEORL-agent starts with the shortest plans but gradually explores longer ones. After around $1000$ episodes, symbolic plans of PEORL-agent converge back to the shortest, indicating that the shortest plans are the overall optimal ones. P-agent also benefits from symbolic plans by not committing improper actions. Furthermore, since ASP-based symbolic planning is usually used to generate shortest plan, P-agent has the steadily largest cumulative reward which happens to be optimal. This result suggests that ASP-based planning can perform very well in deterministic domains where shortest plans are the most desirable. 

In Scenario 2, inspired by \cite[Section 4.1]{option:kulkarni2016intrinsic}, we require that if the taxi arrives at the goal with (4,4) visited, it gets reward 30. The only information present in symbolic knowledge is when (4,4) is visited, the fluent {\em rewardvisited} is set to be true, so that the state representation in RL maps correctly to symbolic space. Again, PEORL-agent outperforms all others (Figure~\ref{fig:taxi-extra-reward}). It starts by trying the shortest plan but during exploration of longer alternatives, it discovers the extra reward, and finally converge to the optimal. Figure~\ref{fig:taxi-route} showed one solution in this scenario. By comparison, since visiting (4,4) is not a necessary condition to drop off the passenger, throughout 10 randomly generated configurations, P-agent never visits that state, behaving the same way with Scenario 1 by sticking to its shortest plan. HRL-agent and RL-agent fail to figure out the extra reward either. This scenario shows that PEORL-agent can discover state with extra reward, and its symbolic plans have leveraged the learned information from RL and become more robust and adaptive to the change of domain details.

\subsection{Grid World}
The Grid World domain is shown in Figure~\ref{fig:gridworld-exp}. Adding to the description in Section \ref{sec:gridworld}, we further assume there are both horizontal and vertical bumpers where the agent receives a penalty of -30 (grids marked as red), and -15 for grids marked with yellow, and -1 for all other grids. Actions {\em grab} and {\em push} have an integer parameter $F$, ranging from 0 to 60, and only if $20\le F <40$ can the execution be successful. Every execution failure causes a -10 penalty. The initial state is chosen from the marked grids in the first column, and goal state is (9,10). We use this example to show that aided by RL, symbolic plans can be learned to avoid bumpers, and can be reliably executed.

We set up RL-agent using Q-learning, PEORL-agent and P-agent. Bumper information is not captured by symbolic knowledge since we assume these are domain details that need to be learned. Learning rates are chosen as the same with Taxi domain. The learning curve is shown in Figure~\ref{fig:gridworld-reward} across 1000 episodes. Similar to the Taxi domain, PEORL-agent has smaller variance in its cumulative rewards (zoomed in by Figure~\ref{fig:gridworld-reward}), and achieves the optimal behavior: it avoids the bumper at its best, and reliably activates and pushes the door (e.g., Figure~\ref{fig:gridworld-exp-route}), surpassing RL-agent. For P-agent, the shortest plans, in this case, are not ideal plans. Since P-agent has no learning capability and only relies on its symbolic knowledge, it performs the worst.

Figure~\ref{fig:gridworld-failures} shows that facing domain uncertainty, the robustness of symbolic plan of PEORL agents is improved using RL, indicated by the reduced number of execution failure. As options mapped from {\em activate} and {\em push} lead to smaller RL problems,
the underlying R-learning quickly learned the right way to execute the options such that the need to replan is significantly reduced. By contrast, relying on replanning, P-agent can recover from failure and eventually achieve its goal, but it cannot improve its execution reliability from learning, leading to poor plan robustness with a relatively large number of execution failures.

\section{Related Work}\label{sec:related}

Integrating symbolic planning and RL has been an active research topic recently. Pre-complied symbolic plans or paths from a finite-state machine play similar roles as options~\cite{parr1998reinforcement,Ryan02usingabstract,leonetti2012automatic}. Recent work also uses ASP to generate longer symbolic plans \cite{leonetti2016synthesis}. In these approaches, symbolic planning is used to help RL through a one-shot plan generation and compilation. By contrast, in our work, planning is interleaved with and constantly updated by RL, and therefore new options can be explored and more meaningful ones will be selected leveraging learning. Our work is also related to automatic option discovery. Various methods have been used, including clustering~\cite{option:mannor2004} and Laplacian Eigenmap~\cite{option:machado2017laplacian}. To the best of our knowledge, this is the first work using symbolic planning for automatic option discovery. Our work is also motivated by improving symbolic planning through learning. Different learning models have been adopted in earlier work such as relational decision tree \cite{COIN:COIN447} or weighted exponential average \cite{kha14}. RL is also used to improve task decomposition in HTN planning \cite{hogg2010learning}. Finally, integrating planning with probabilistic planning on POMDP was investigated \cite{zhang2017dynamically}.


\section{Conclusion and Future Work}\label{sec:con}
In this paper we proposed the PEORL framework, where symbolic planning and HRL simultaneously improve each other, leading to rapid policy search and robust symbolic planning. Future work involves formal study on hierarchical R-learning, using PEORL framework to solve more complicated domains, and integrating symbolic planning with deep RL for interpretable end-to-end solutions.

\section*{Acknowledgements}
Bo Liu and Daoming Lyu are funded by the startup funding at Auburn University.

\newpage

\bibliographystyle{named}
\bibliography{bib}

\end{document}